%% file: _main.tex
\documentclass[letterpaper, 10 pt, conference]{formatting/ieeeconf}
\IEEEoverridecommandlockouts                              
\let\labelindent\relax
\usepackage{amsfonts}       

\usepackage{amsthm}
\usepackage{wrapfig}
\usepackage{svg}
\usepackage{mathtools}      
\usepackage{amssymb}        

\usepackage{filecontents}
\usepackage{graphicx}       
\usepackage{marginnote}     
\usepackage{marvosym}       
\usepackage{overpic}        
\usepackage{tabularx}
\usepackage{cite}
\usepackage{color}
\usepackage[linesnumbered,algoruled,boxed,lined]{algorithm2e}
\usepackage[normalem]{ulem}
\usepackage{epstopdf}
\usepackage{float}
\usepackage{enumitem}
\usepackage[normalem]{ulem}
\usepackage{comment}
\usepackage[font=footnotesize,labelfont=bf]{caption}    
\usepackage[a-2b,mathxmp]{pdfx}[2018/12/22]

\newif\ifdraft
\draftfalse

\ifdraft
\usepackage[paperheight=11in,paperwidth=9.5in,
			left=1.25in,right=1.25in,
			top=0.75in,bottom=0.75in,
			heightrounded,marginparwidth=1.2in,
			marginparsep=0.05in]{geometry}
\usepackage{xcolor}
\usepackage{xargs} 
\usepackage[textsize=footnotesize]{todonotes}
\newcommandx{\nt}[2][1=]{\todo[linecolor=red,
			backgroundcolor=red!10,bordercolor=red,#1]{ #2}}
\newcommandx{\jy}[2][1=]{\todo[linecolor=green,
			backgroundcolor=green!10,bordercolor=green,#1]{JY: #2}}
\else
\newcommand{\nt}[1]{{}}
\newcommand{\jy}[1]{{}}
\fi

\newif\iftwocolumn
\twocolumntrue

\theoremstyle{definition}

\theoremstyle{remark}

\SetKwProg{Fn}{Function}{}{}
\SetKwComment{Comment}{$\triangleright$\ }{}

\makeatletter
\def\subsubsection{\@startsection{subsubsection}
                                 {3}
                                 {\z@ \hspace*{1mm}}
                                 {0ex plus 0.1ex minus 0.1ex}
                                 {0ex}
                                 {\normalfont\normalsize\itshape}}
\makeatother

\newcommand{\nlra}{\textsc{SSRA}\xspace} 
\newcommand{\glra}{\textsc{GSSA}\xspace}  
\newcommand{\lwlr}{\textsc{LWLR}\xspace} 
\newcommand{\damone}{\textsc{DAM}\xspace}
\newcommand{\damtwo}{\textsc{DAMv2}\xspace}
\newcommand{\metricddd}{\textsc{M3Dv2}\xspace}

\newcommand{\mdem}{\textsc{MDEM}\xspace}
\newcommand{\moma}{\textsc{MOMA}\xspace}

\font\titlefont=ptmb at 14.9pt
\title{\titlefont
Monocular One-Shot Metric-Depth Alignment for RGB-Based Robot Grasping
}
\author{Teng Guo \qquad Baichuan Huang     \qquad Jingjin Yu
\thanks{G. Teng, B. Huang  and J. Yu are with the Department of 
Computer Science, Rutgers University, Piscataway, NJ, USA. 
Emails: {\tt\small \{teng.guo, baichuan.huang, jingjin.yu\}@rutgers.edu}. This work was supported in part by NSF awards IIS-1845888, IIS-2132972, and CCF-2309866.
}
}

\begin{document}

\maketitle
\thispagestyle{empty}
\pagestyle{empty}


\begin{abstract}
Accurate 6D object pose estimation is a prerequisite for successfully completing robotic prehensile and non-prehensile manipulation tasks. 
At present, 6D pose estimation for robotic manipulation generally relies on depth sensors based on, e.g., structured light, time-of-flight, and stereo-vision, which can be expensive, produce noisy output (as compared with RGB cameras), and fail to handle transparent objects. 
%
On the other hand, state-of-the-art monocular depth estimation models (\mdem{s}) provide only affine-invariant depths up to an unknown scale and shift. Metric \mdem{s} achieve some successful zero-shot results on public datasets, but fail to generalize.   
We propose a novel framework, \emph{monocular one-shot metric-depth alignment}, \moma, to recover metric depth from a single RGB image, through a one-shot adaptation building on \mdem techniques. 
%
\moma performs scale-rotation-shift alignments during camera calibration, guided by sparse ground-truth depth points, enabling accurate depth estimation without additional data collection or model retraining on the testing setup. \moma supports fine-tuning the \mdem on transparent objects, demonstrating strong generalization capabilities. Real-world experiments on tabletop 2-finger grasping and suction-based bin-picking applications show \moma achieves high success rates in diverse tasks, confirming its effectiveness. 
\end{abstract}

\section{Introduction}\label{sec:intro}
\input{texs/00-intro}


\section{Preliminaries: Metric Depth Recovery from Affine-Invariant \mdem}\label{sec:problem}
\input{texs/02-prelim}

\section{Monocular One-Shot Metric-Depth Alignment}\label{sec:algorithm}
\input{texs/03-methods}

\section{Performance Evaluation}\label{sec:evaluation}
\input{texs/06-evaluation}

\section{Conclusion and Discussions}\label{sec:conclusion}
\input{texs/07-conclusion.tex}

\bibliographystyle{formatting/IEEEtran}
\bibliography{bib/all}

\end{document}

%% file: texs/00-intro.tex
Fulfilling the future promise of robotics technology, be it doing repetitive work at warehouses, serving patients and nurses at hospitals, or helping with chores in our homes, critically depends on enabling robots to agilely and robustly manipulate a wide variety of objects. 
Skillful robotic manipulation, in turn, demands a sufficiently accurate understanding of the scene including the 6D poses of the target object and its surroundings~\cite{fang2023anygrasp, cao2021suctionnet, sundermeyer2021contact, mousavian20196}. 
Presently, pose estimation for robotic manipulation tasks, such as grasping, resorts to the use of depth sensors based on various hardware-software stacks, e.g., structured light, time-of-flight distance measurement, stereo-vision, and so on. 
Despite often carrying hefty price tags\footnote{An RGB-D sensor costs anywhere between 500 to 20K+ USD.}, output from these depth sensors can be frequently noisy with wrong and missing information. Factors such as transparent objects and light interference present further challenges for depth sensors. 
Issues such as steeper costs and reliability concerns limit the development and application of depth sensor-based robotic manipulation, which raises the question: can we perform 6D pose estimation without resorting to constantly using depth sensors?
\begin{figure}[!htbp]
    \centering
    \includegraphics[width=1\linewidth]{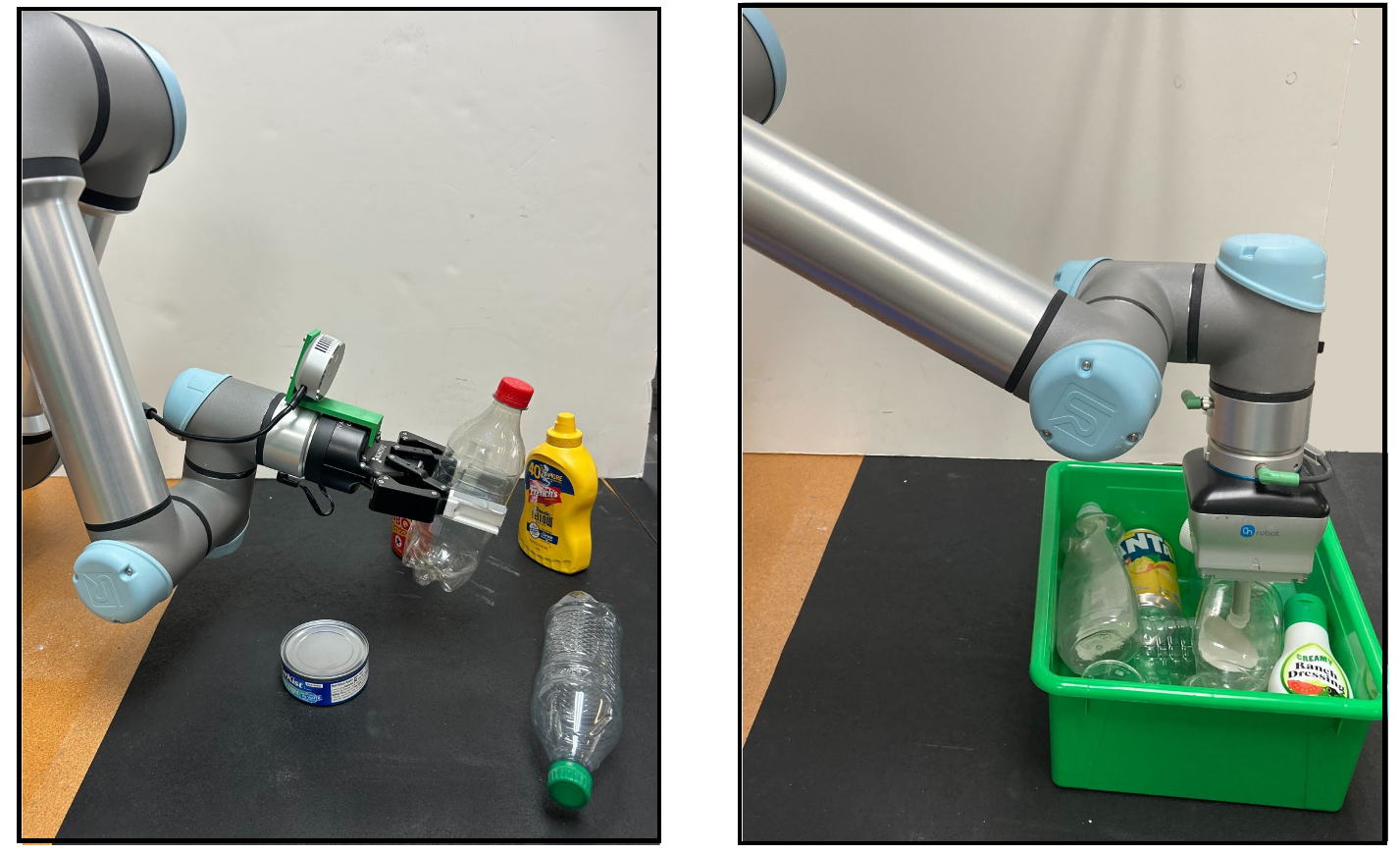}
    \caption{Two downstream applications over which our framework, Monocular One-shot Metric-depth Alignment (\moma), was tested against: two-finger grasping on tabletop setting and suction-based bin-picking. Using only RGB image, \moma enable the robot successfully pick challenging transparent objects in cluttered scences.}
    \label{fig:robots}
\end{figure}

\begin{figure*}[!htbp]
    \centering
    \includegraphics[width=1\linewidth]{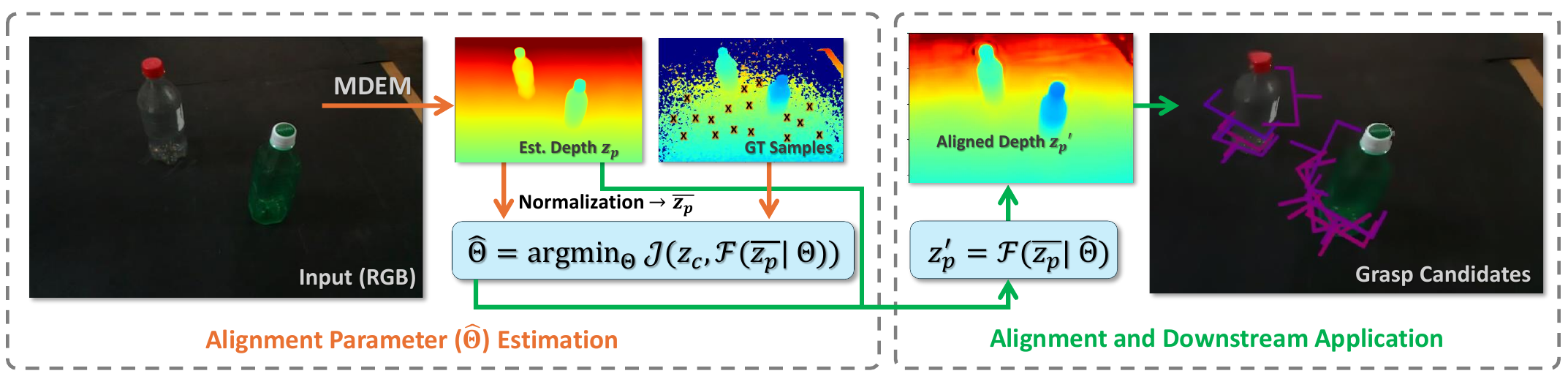}
    \caption{The overall operating pipeline of \moma. From \mdem, raw estimated depth is obtained and normalized to obtain normalized depth $\overline{z_p}$. For a new camera/environment setup, a one-time alignment parameter estimation is performed to obtain $\hat{\Theta}$ using $\overline{z_p}$ and some ground truth depth points. $\hat{\Theta}$ can then be used for calibrate \mdem output to produced aligned depth $z_p'$ for downstream applications. Note: to simplify the illustration, we used a single scene for both parameter estimation and alignment; in practice, scenes in applications are generally not the same as those used for parameter estimation..}
    \label{fig:framework}
\end{figure*}
A simple experiment suggests this should be possible through a data-driven approach: most humans can easily grasp and manipulate objects, even with an eye closed (i.e., no stereo vision). 
In other words, based on monocular vision and experience, humans can accurately estimate depth information for common manipulation tasks. 
Indeed, prior research, including state-of-the-art transferable methods such as MiDaS \cite{birkl2023midas} and LeReS \cite{yin2021learning}, made concrete attempts to estimate depth information based on a single image through learning. At the same time, these methods output affine-invariant depth estimates, which are accurate up to an \emph{unknown} offset and scale, making them unsuitable for physical interaction/manipulation tasks. 

Toward developing general machine learning-based methods for enabling robust pose estimation for robotic manipulation, in this work, we focus on the setting where \emph{camera poses are fixed}, which is the dominant lab setting for fixed robot arms and the setting in most industrial warehouse applications, e.g., bin-picking. 
We introduce a novel yet simple-to-implement framework, \emph{monocular one-shot metric-depth alignment} (\moma), to recover metric depth from the output of state-of-the-art \emph{monocular depth estimation models} (\mdem), e.g., \damone~\cite{Yang2024DepthAU}. \moma achieves this through carefully designed scale-rotation-shift alignment during the camera calibration phase, guided by a sparse set of ground-truth depth points. Afterward, our robotic system can accurately estimate depth from single RGB images. 
%


\moma is a one-shot approach that eliminates the need to collect data in the target environment and retrain \mdem models. Through fine-tuning \damone on transparent object datasets, \moma readily \emph{generalizes} to these objects as well. Due to its design, \moma is also very fast, taking only a few seconds to perform an alignment calibration and a few milliseconds during runtime to process an \mdem output. The effectiveness of \moma is thoroughly benchmarked and validated in experiments using a UR-5e manipulator, achieving remarkable success rates in both two-finger grasping tasks on tabletop settings and suction-based bin-picking tasks. The diverse evaluations demonstrate the promise of \moma as a low-cost, depth-sensor-free depth-estimation approach for real-world manipulation tasks.


\textbf{Organization.} 
%
Sec.~\ref{sec:related} discusses related work. 
In Sec.~\ref{sec:problem}, we introduce the key alignment problem and relevant solution approaches. 
Then, in Sec.~\ref{sec:algorithm}, we present our pipeline that uses \mdem and scale-rotation-shift depth alignment module.
In Sec.~\ref{sec:evaluation}, we evaluate the effectiveness of our algorithms on various maps. 
We conclude and discuss future directions for research in Sec.~\ref{sec:conclusion}.

\section{Related Work}\label{sec:related}
\textbf{Depth Estimation}. Depth-estimation methods can be classified into three categories based on whether they learn metric depth, relative depth, or affine-invariant depth. Metric depth estimation aims at providing accurate (Euclidean) depth estimation. Whereas existing methods~\cite{bhat2023zoedepth, bhat2021adabins, Yang2024DepthAU, yang2024depthv2} have achieved impressive accuracy, they are limited to function within datasets coupled with fixed camera intrinsics. This results in training datasets for metric depth methods often being small, since it is challenging to collect large datasets covering diverse scenes using a single camera. Consequently, the trained models are not transferable and generalize poorly to images in unseen scenarios, esp., when camera parameters of test images differ. A compromise is to learn relative depth~\cite{xian2018monocular, xian2020structure, chen2020oasis, chen2016single}, which only indicates whether one point is farther or closer than another. This type of depth estimate, while qualitative useful, is unsuitable for applications such as manipulation. Learning affine-invariant depth~\cite{birkl2023midas, yin2023metric3d, yang2024depthv2, Yang2024DepthAU} offers a trade-off between metric and relative depth. With large-scale data, affine-invariant approaches decouple the metric information during training and achieve impressive robustness and generalization capability. All existing methods require collecting new data and model retraining for a new target camera/application setup.

Focusing on \emph{monocular (metric) depth estimation}, the topic of this work, deep-learning-based methods dominate, which learn depth representations from delicately annotated datasets \cite{geiger2013vision,silberman2012indoor}.
Recently, numerous new models have emerged \cite{bhat2021adabins, yin2023metric3d, yang2024depth, yang2024depthv2, bhat2023zoedepth, birkl2023midas, yin2021learning} that can handle open-world images. While these models claim to have a zero-shot generalization capability in terms of metric-depth estimation, the predicted depth is insufficiently precise for tasks involving physical interaction, e.g., robotic manipulation.

\textbf{Grasp Pose Estimation}.
Given a scene with object pose estimations, \emph{grasp pose estimation} proposes poses for end-effectors to grasp target objects. 
%
%
Grasp pose estimation turns out to be a task well-suited for data-driven approaches \cite{fang2023anygrasp, fang2020graspnet, cao2021suctionnet, sundermeyer2021contact, mousavian20196}. While depth sensors may struggle with transparent objects, depth-completion methods such as those in \cite{sajjan2020clear,fang2022transcg,shi2024asgrasp} have been proposed to obtain more accurate depth information.
A recent representative work, MonoGraspNet~\cite{zhai2023monograspnet}, aims to generate grasp poses and regress depth using a single image. The method faces generalization challenges, as is the case for other \mdem models, to diverse scenes and different camera intrinsics/poses. To generalize to other scenarios and camera setups, one has to collect a new dataset and train for their specific settings which can be tedious and limits the application of these methods.
In addition to single-view approaches, multi-view NeRF-based approaches \cite{dai2023graspnerf, kerr2022evo, liu2024rgbgrasp} have been explored for object reconstruction and grasp pose prediction. These methods, requiring non-trivial online computation time, provide an alternative depth sensor-free estimation methods for metric depth. 
%


%% file: texs/02-prelim.tex
\subsection{The Metric Depth Alignment Problem}
Metric monocular depth estimation from a single image is inherently ill-posed as a single 2D image lacks the necessary information to uniquely determine the 3D distances and scale of objects without additional context or assumptions~\cite{yin2023metric3d}.
Some monocular depth estimators focus on evaluating relative (scale-and-shift-invariant or affine-invariant) depth, thereby enhancing generalization capabilities from a single image. Several vision foundation models have been developed for metric depth estimation, demonstrating admirable performance on existing public datasets. 
However, due to insufficient \emph{metric depth} accuracy at the instance level, these models remain impractical for robotics applications.
%

Let $\mathbf{z}_c \in \mathbb{R}^{1\times n}$ be $n$ sampled ground truth depth points in camera coordinate system, $\mathbf{z}_p \in \mathbb{R}^{1\times n}$ the corresponding predicted depth from the depth estimation model, and $\mathbf{u} \in \mathbb{R}^{1\times n}$ and $\mathbf{v} \in \mathbb{R}^{1\times n}$ the corresponding pixel row/column indices.
The metric depth alignment problem can be formulated as the following optimization problem:

\begin{equation}\label{eq:prob}
    \min_{\Theta} \mathcal{J}\left(\mathbf{z}_c, \mathcal{F}(\mathbf{u}, \mathbf{v}, \mathbf{z}_p \mid \Theta)\right),
\end{equation}
with $\mathcal{J}$ being a suitable cost function, $\mathcal{F}$ the mapping function from predicted depth to ground truth depth, and $\Theta$ the parameters.


\subsection{Prior Alignment Methods}
While not aiming for fine-granularity downstream applications, prior works have tackled the alignment problem by performing \emph{linear} transformation to adjust scale and shift (translate) of depth via solving

\begin{equation}\label{eq:prob}
    \min_{s,t} \parallel \mathbf{z}_c - (s \mathbf{z}_p + t) \parallel^2
\end{equation}
where $s$ and $t$ are scaling and shift scalars, respectively. 

The problem described in Eq.~\eqref{eq:prob} can be solved using a global least-squares fitting method \cite{yin2021learning,ranftl2020towards} which we denote as \emph{global scale-shift alignment} (\glra). Such methods use a single set of scale/shift scalars, which cannot accommodate spatial heterogeneity. \emph{locally weighted linear regression} (\lwlr) \cite{xu2024toward} proposes to use sparse ground-truth depth to compute a scale-shift recover map, which solves a weighted linear regression problem for each pixel $(i,j)$:

\begin{equation}
\begin{aligned}
      &\min_{\beta_{ij}}(\mathbf{z}_c - \beta_{ij} \mathbf{Z}_p)^{\top} \mathbf{W}_{ij} (\mathbf{z}_c - \beta_{ij} \mathbf{Z}_p) \\
      &\mathbf{Z}_{p}=[\mathbf{z}_p, 1]^{\top}, \mathbf{W}_{ij}=\text{diag}(w_1,\ldots,w_k,\ldots,w_n)\\
      &\beta_{ij}=[s_{ij},t_{ij}]^{\top} \in \mathbb{R}^{2\times 1} \\
      &\hat{\beta}_{ij}=[\hat{s}_{ij},\hat{t}_{ij}]^{\top}=(\mathbf{Z}_p^{\top}\mathbf{W}_{ij}\mathbf{Z}_p)^{-1}\mathbf{Z}_p^{\top}\mathbf{W}_{ij}\mathbf{z}_c \\
      &\hat{\mathbf{D}}_p=\mathbf{S}\odot \mathbf{D}_p+\mathbf{T}, \mathbf{S}_{ij}=\hat{s}_{ij}, \mathbf{T}_{ij}=\hat{t}_{ij}
\end{aligned}
\end{equation}
where $w_k = \frac{1}{\sqrt{2\pi}} e^{-d_k^2/(2b^2)}$, $d_k$ is the Euclidean distance of the $k$-th pixel to the pixel $(i,j)$,  $b$ is a predefined bandwidth of the Gaussian kernel and $\hat{\mathbf{D}}_p$ is the final aligned depth from the predicted depth of \mdem. 
%
Whereas \lwlr improves over global scale/shift alignment methods, its output remains insufficiently accurate for manipulation tasks.

%% file: texs/03-methods.tex
\subsection{Overview of \moma}
As illustrated in Fig.~\ref{fig:framework}, \moma leverages \mdem{}s, often trained on gigantic data sets, to estimate the depth of an object from an  RGB image, which ensures broad object coverage. 
\moma then compute parameters $\hat{\Theta}$ (exact form of $\Theta$ to follow) in~Eq.~\eqref{eq:prob} through solving a \emph{non-linear} optimization problem aligning a sparse set of ground truth depth information (easily obtainable for cameras with known poses with respect to the scene or with an aligned depth camera) with the corresponding points from the \mdem output. 
For downstream tasks, monocular RGB images are processed using \mdem and then aligned using the obtained parameters $\hat{\Theta}$. 
The camera pose is assumed to remain unchanged, which corresponds to  the majority of fixed-arm use cases in research labs and in industrial applications. If camera poses change, quick re-calibrations can be performed.
%


\subsection{Scale-Shift-Rotation Alignment of \mdem Output}
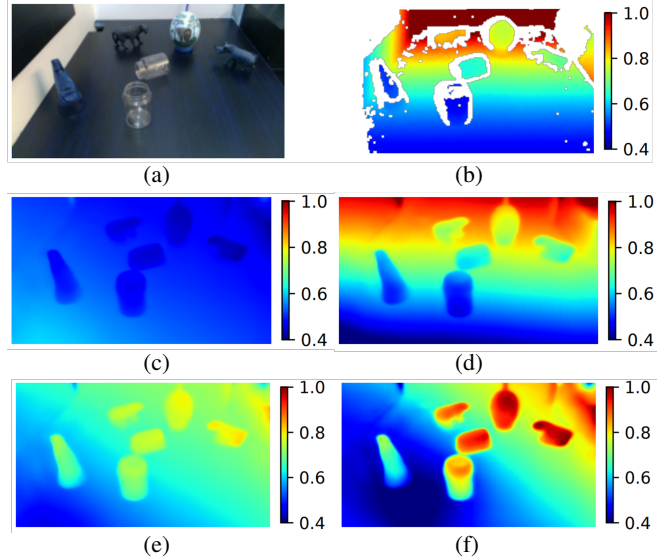
\begin{figure}[tb]
    \centering
        
 \begin{overpic}[width=1\linewidth]{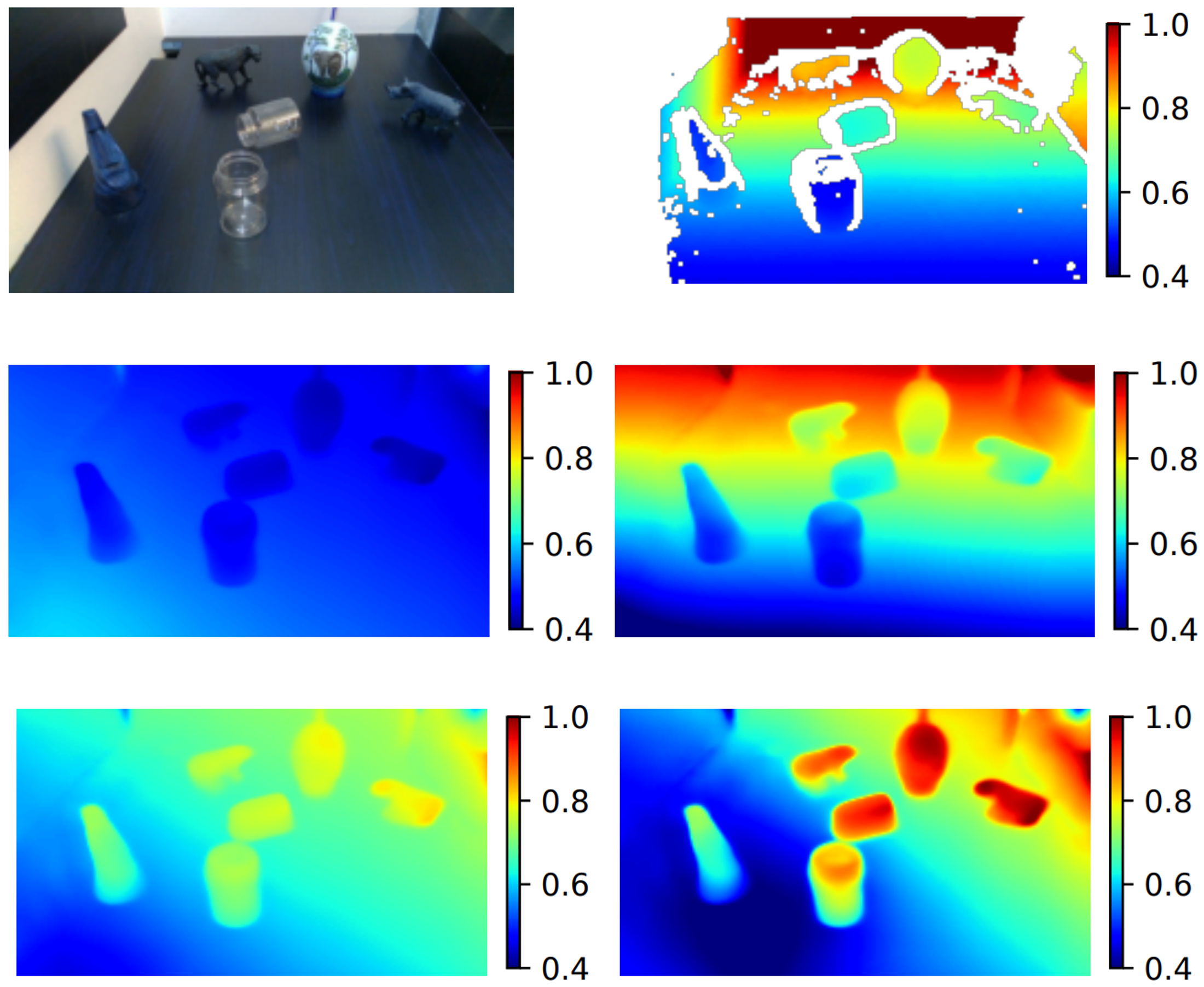}
             \small
             \put(21, 54) {(a)}
             \put(69, 54) {(b)}
             \put(21, 25) {(c)}
             \put(69, 25) {(d)}
              \put(21, -3) {(e)}
           \put(69, -3) {(f)}
        \end{overpic}
    \vspace{-1mm}
    \caption{An extreme example case showcasing the capability of \nlra. (a) The RGB image. (b) The ground-truth depth.  (c) The predicted depth from \damone fine-tuned on TransCG~\cite{fang2022transcg}; the input scenario could be treated entirely unseen by the \mdem, resulting in low-quality depth output. (d) Aligned depth using \nlra, which yields better results compared to \lwlr and \glra. (e) Aligned depth using \glra.  (f) Aligned depth using \lwlr.
}
    \label{fig:scaleshift}
\end{figure}

Whereas \mdem{}s are not new, they only start to approach foundation model level of quality recently \cite{Yang2024DepthAU,yang2024depthv2}. These newer models, trained over very large data sets, are capable of producing decent depth estimates for a very diverse set of scenes. 
%
Careful examination of the zero-shot capability of these vision foundation models, however, reveals situations where haywire predictions are often made. An extreme example can be seen in Fig.~\ref{fig:scaleshift}(c), where the \mdem predicted depth completely reverses the relative depth information.

Feeding more data to \mdem{}s can help, but there is no guarantee hallucinations can be eliminated when a previously unknown scene is encountered. On the other hand, we observe that it does not require huge efforts to adapt a capable \mdem model to a new target camera/environment, leading us to propose a \emph{targeted} depth recovery method that considers \emph{scale}, \emph{shift}, and \emph{rotation} factors.

To calibrate the \mdem output for a target camera/environment, we first capture an RGB image $\mathcal{I}$ and obtain its corresponding sparse ground-truth depth in the camera coordinate system.  The ground-truth depth generally only requires tens to hundreds of sample points. 
%
%
Let $\mathbf{X}_{pk} = [x_{pk}, y_{pk}, z_{pk}]$ be the 3D point for the $k$-th pixel predicted by the model, and $\mathbf{X}_{ck} = [x_{ck}, y_{ck}, z_{ck}]$ be the corresponding 3D point in the camera coordinate system.
We align these two point clouds by finding $s\in \mathbb{R}, \mathbf{R}\in SO(3)$ and $\mathbf{T}\in \mathbb{R}^{1\times 3}$ to minimize 
\begin{equation}\label{eq:pd_align}
    \frac{1}{n}\sum_{k=1}^{n}\| s\mathbf{R} \mathbf{X}_{pk} + \mathbf{T} - \mathbf{X}_{ck} \|^2
\end{equation}



Assuming pinhole camera models, let $c_{xp}$, $c_{yp}$, and $f_p$ be the (pseudo) intrinsics for the \mdem. In other words, the depth prediction of \mdem is treated as a depth map captured by a pseudo depth sensor with the defined intrinsic. We have

\begin{equation}
x_{pk} = \frac{z_{pk}(u_k - c_{xp})}{f_p}, \quad y_{pk} = \frac{z_{pk}(v_k - c_{yp})}{f_p}
\end{equation}

Since we are interested only in depth, it is unnecessary to estimate all parameters listed in Eq.~\eqref{eq:pd_align}. Instead, we focus solely on minimizing the fitted depth error. In other words, we consider only the minimizing the depth alignment error

\begin{equation}\label{eq:nonlinear}
      \min_{\Theta}\frac{1}{n}\sum_{k=1}^{n}\parallel z_{ck}- \mathcal{F}(u_k,v_k,z_{pk}|\Theta) \parallel^2 
\end{equation}
where 

\begin{equation}
\begin{gathered}
      \mathcal{F}(u_k,v_k,z_{pk}|\Theta) =s(R_{31}x_{pk}+R_{32}y_{pk}+R_{33}z_{pk})+T_3\\
= s( - x_{pk} \sin\phi+ y_{pk} \sin\theta \cos\phi+z_{pk} \cos\theta \cos\phi  )+T_3\\
\end{gathered}
\end{equation}
and

\begin{equation}
            \Theta = [s,\theta, \phi, T_3, c_{xp}, c_{yp}, f_p]^\top.
\end{equation}


Solving the non-linear fitting problem in Eq.~\eqref{eq:nonlinear} yields optimized parameters $\hat{\Theta}$ (which is a subset of all parameters from~Eq.~\eqref{eq:pd_align}). 
We denote this method as \emph{scale-shift-rotation alignment} or \nlra, which is a specific realization of \moma. 
Alignment using only scale and shift can be seen as a special case where $\theta=\phi=0.$

\subsection{One-Shot Calibration and Depth Normalization} 
To estimate $\hat{\Theta}$, it is necessary to have some form of ground truth depth information. Here, we assume that a portion of the scene will remain the same for a fixed camera/environment setup, especially in robotic manipulation tasks. For example, in a tabletop setting, part of the table surface will remain visible across different scenes. Therefore, if we sample a sparse, diverse set of ground truth points, they can serve as reliable references. 

\begin{figure}[t!]
    \centering
    \includegraphics[width=1\linewidth]{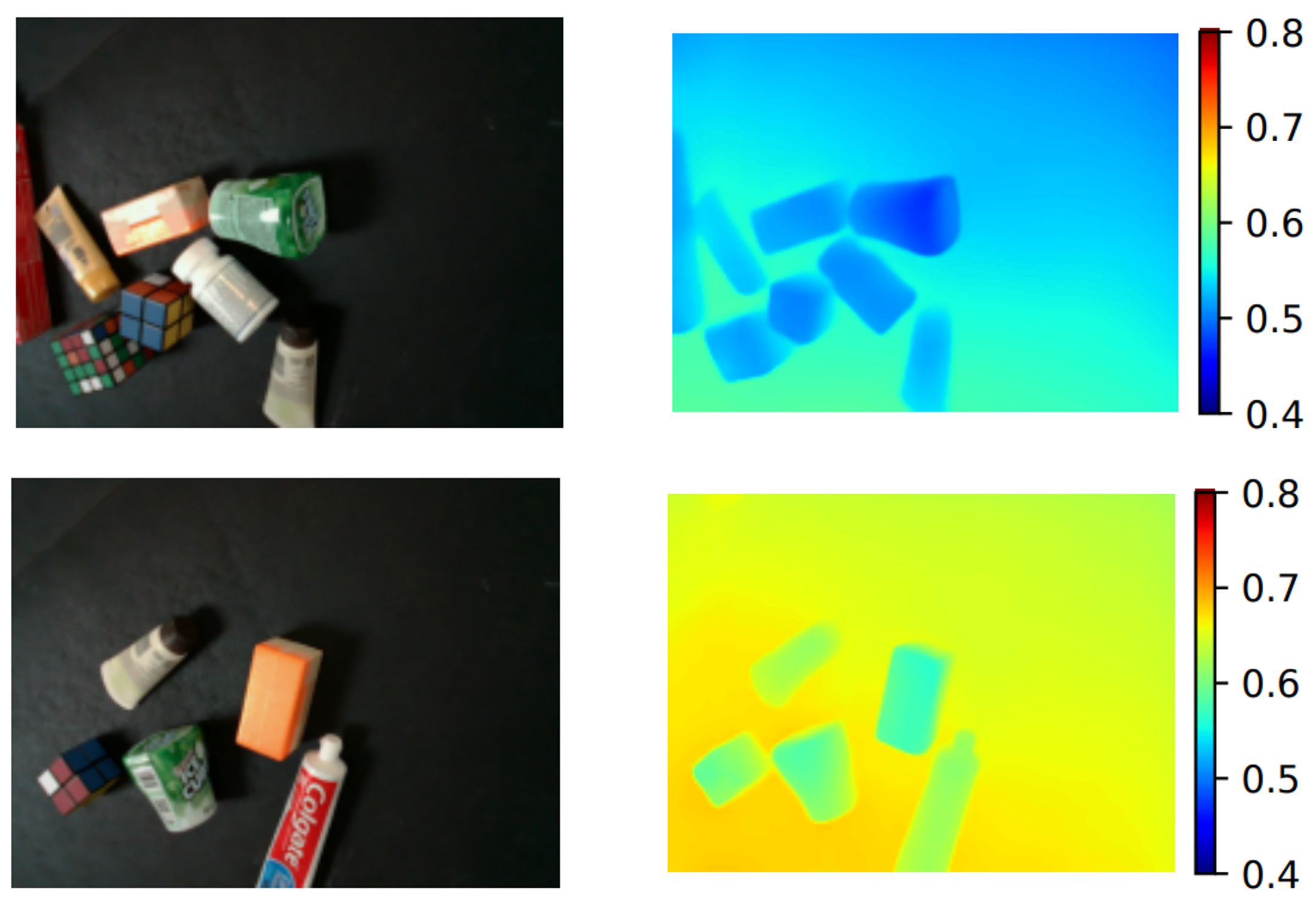}
    \caption{An example showing \mdem prediction fluctuations. The top and bottom RGB images on the left have the same fixed camera/background. However, the \mdem depth outputs (shown on the right) vary drastically.
 }
    \label{fig:fluctuate}
\end{figure}

However, there is a challenge. The raw predicted depth of \mdem will fluctuate when different objects are placed in the scene, even though the camera pose is fixed. Consequently, \mdem predicted depth at the same location will change. This renders one-shot calibration results invalid. An example given in in Fig.~\ref{fig:fluctuate}.
We normalize the predicted depth $\mathbf{z}_p$ to minimize the issue's impact. We examined two normalization methods, with the first being \cite{ranftl2020towards}:

\begin{equation}
    m(\mathbf{z}_p) = \text{median}(\mathbf{z}_p), \quad \mu(\mathbf{z}_p) = \frac{1}{n} \sum_{i=1}^{n} |\mathbf{z}_p - m(\mathbf{z}_p)|
\end{equation}

\begin{equation}
    \hat{\mathbf{z}}_p = \frac{\mathbf{z}_p - m(\mathbf{z}_p)}{\mu(\mathbf{z}_p)}
\end{equation}

The second is min-max normalization~\cite{garcia2015data}:

\begin{equation}
    z_{\min} = \min(\mathbf{z}_p), \quad z_{\max} = \max(\mathbf{z}_p)
\end{equation}

\begin{equation}
    \hat{\mathbf{z}}_p = \frac{\mathbf{z}_p - z_{\min}}{z_{\max} - z_{\min}} + z_{\min}
\end{equation}

Experiments show the second approach leads to better results.
After obtaining the normalized depth $\hat{\mathbf{z}}_p$, we substitute it into Eq.~\eqref{eq:nonlinear} to compute the parameters $\hat{\Theta}$. When performing inference, the predicted depth output by \mdem will be normalized similarly, and the final depth is computed using $\mathcal{F}(\cdot)$ from Eq.~\eqref{eq:nonlinear}.

%% file: texs/06-evaluation.tex
We evaluate the proposed method using standard depth estimation metrics. All metrics are calculated within object areas defined by object masks unless specified otherwise. The primary \mdem used in our evaluation is \damone~\cite{Yang2024DepthAU} ViT-L model. The metrics include:

\begin{itemize}
    \item \textbf{RMSE}: The root mean squared error (in meters) between the depth estimates and the ground-truth depths.
    \item \textbf{REL}: The mean absolute relative difference between the estimated and ground-truth depths.
    \item \textbf{MAE}: The mean absolute error (in meters) between the depth estimates and the ground-truth depths.
    \item \textbf{Threshold $\delta$}: The percentage of pixels for which the predicted depths satisfy $\max\left(\frac{d}{d_p}, \frac{d_p}{d}\right) < \delta$, where $d$ and $d_p$ represent corresponding pixels from the depth maps $D$ and $D_p$, respectively. The thresholds $\delta$ are commonly set to 1.05, 1.10, and 1.25.
\end{itemize}

The pre-trained \damone underperforms on transparent objects; we fine-tuned it on transparent object datasets, combining around 500K images from \cite{sajjan2020clear, chen2022clearpose, fang2022transcg, zhu2021rgb}. The training was performed with a learning rate of $5\times 10^{-4}$, a weight decay of 0.01, and a LoRA~\cite{hu2021lora} rank of 256 over 20 epochs. The scale-invariant loss from \cite{bhat2022localbins} was utilized for training. 
For \glra, $b=100$ is used in the evaluation.
On a PC equipped with Nvidia RTX 3090, \damone takes about 0.6 seconds to process an input monocular image. Performing alignment at run time (processing \damone output to obtain metric depth) takes less than 10 milliseconds. Running time for the one-shot alignment parameter calibration will be presented in Sec.~\ref{sec:n}.
The code and tested dataset can be found in \href{https://github.com/GreatenAnoymous/MonoDPT_grasp}{https://github.com/GreatenAnoymous/MonoDPT$\_$grasp}.

\subsection{Key Performance Metrics under Multiple Camera Poses}\label{subsec:poses}
\begin{figure}[t!]
    \centering
        \includegraphics[width=1\linewidth]{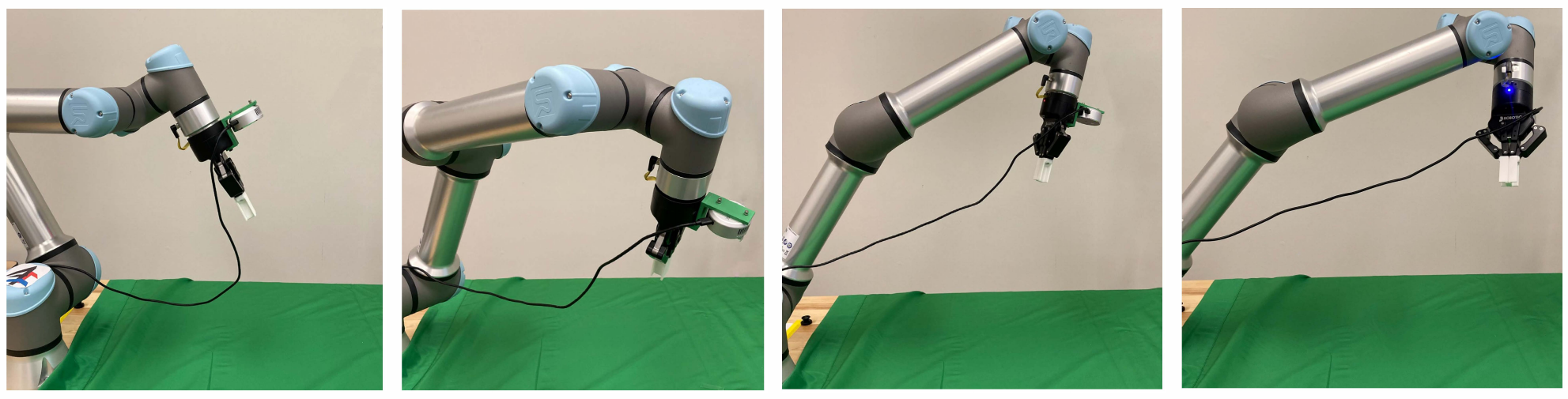} \vspace{-2mm}\\
\begin{overpic}          
        [width=1\linewidth]{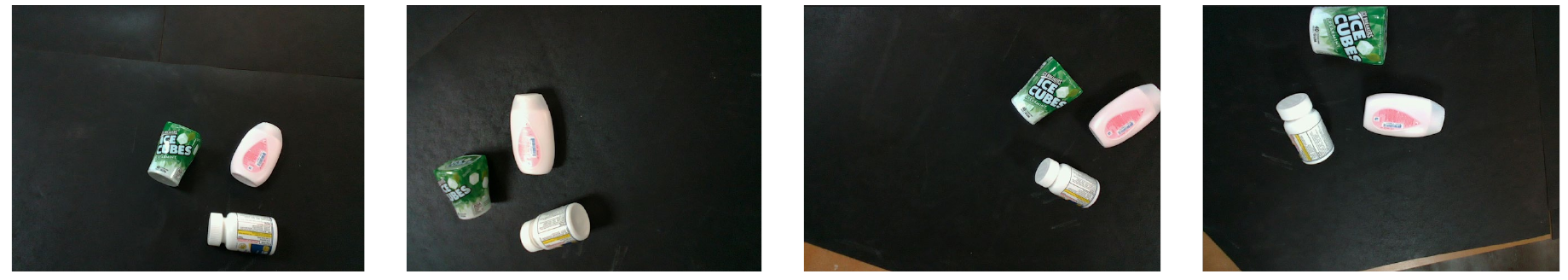}
             \small
             \put(9.0, -3) {(1)}
             \put(34.5, -3) {(2)}
             \put(59.5, -3) {(3)}
             \put(85.5, -3) {(4)}
        \end{overpic}
    \vspace{0mm}
    \caption{[top] Four diverse camera poses used in our evaluation. [bottom] Corresponding sample scenes taken at the four poses.}
    \label{fig:pose_study}
\end{figure}
To evaluate performance, we first use a Realsense L515 to collect a dataset for 20 opaque objects under four fixed camera poses. For each camera pose, we collect 100 RGB/D images and generate the object masks using SAM~\cite{kirillov2023segment}. 
Since \damone\ is not trained on any images from our setup, these scenarios are considered unseen to \damone.
When evaluating the methods on the dataset for a given pose, we first use an RGB/D image to compute alignment parameters $\hat{\Theta}$, using 100 samples from the depth image. The parameters $\hat{\Theta}$ are used for the rest of the evaluations. 

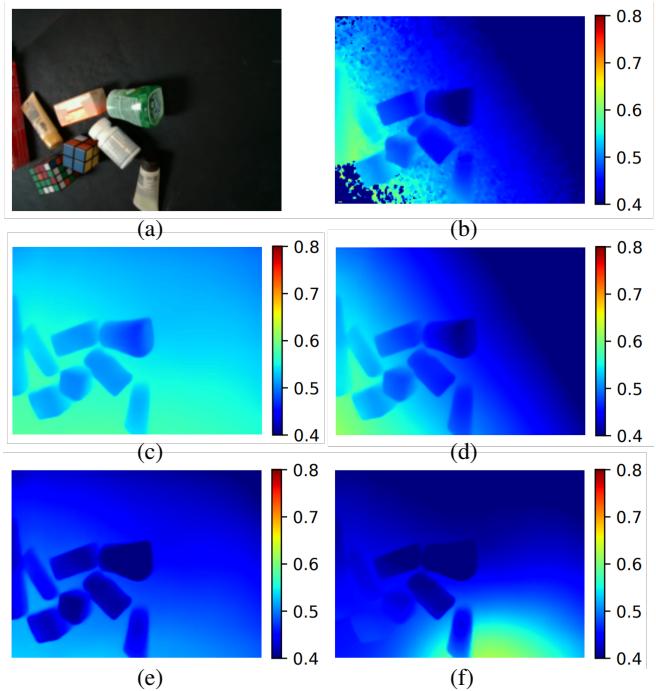
\begin{figure}[t!]
    \centering
 \begin{overpic}[width=1\linewidth]{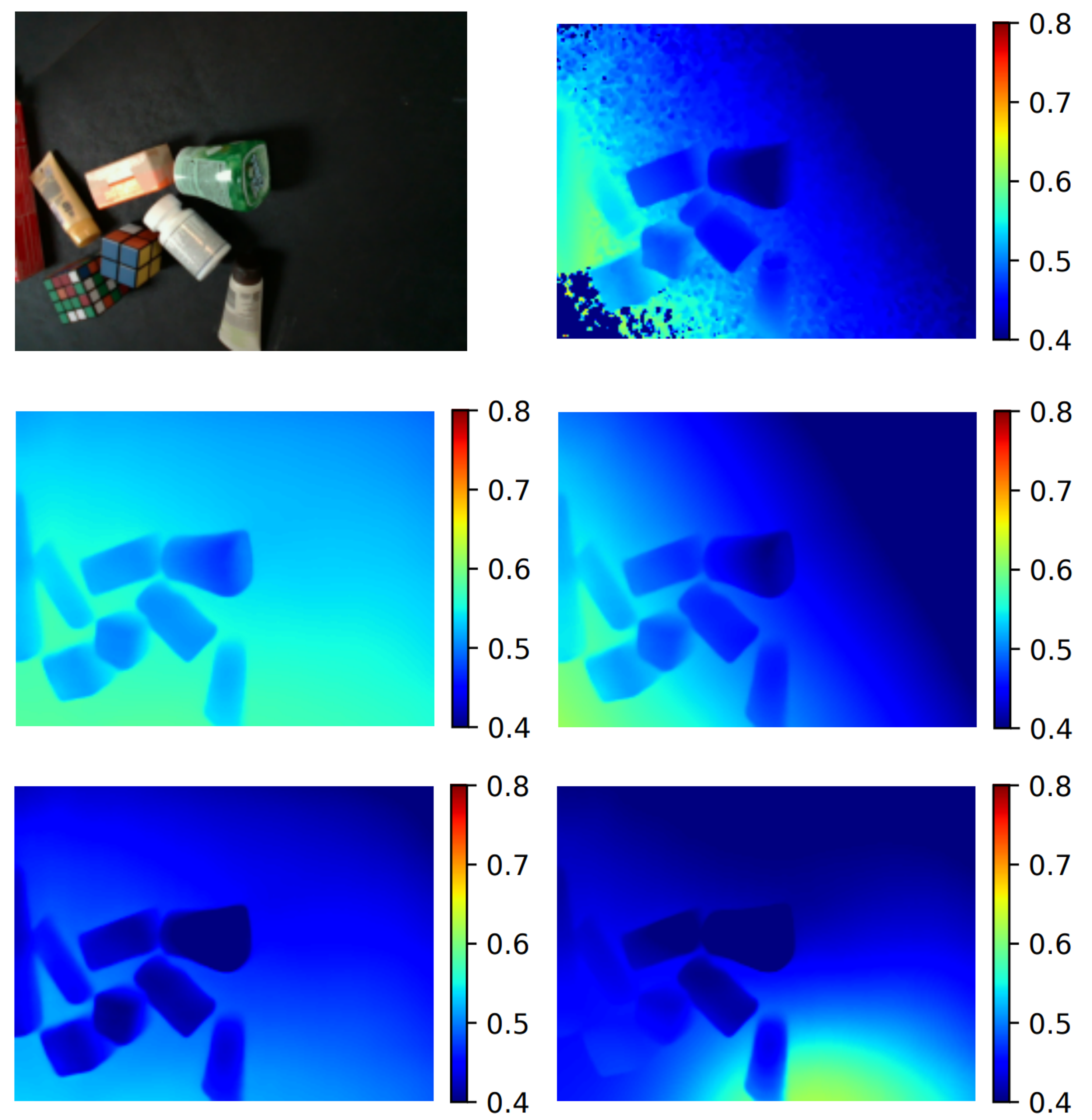}
             \small
             \put(20, 65) {(a)}
             \put(67, 65) {(b)}
             \put(20, 31.5) {(c)}
             \put(67, 31.5) {(d)}
              \put(20, -2.5) {(e)}
           \put(67, -2.5) {(f)}
        \end{overpic}
    \vspace{-1mm}
    \caption{(a) The RGB image. (b) The ground-truth depth. (c) The predicted depth from \damone;(d) Aligned depth using \nlra, which yields better results compared to \lwlr and \glra.   (e) Aligned depth using \glra.  (f) Aligned depth using \lwlr.
}
    \label{fig: exampleresult}
\end{figure}

Experimental results in Tab.~\ref{t:t1} show the performance of four different methods (\nlra, \glra, \lwlr, and \damone raw metric depth output without any alignment) across four different poses for opaque objects. Best performances are highlighted using bold font. An example containing output from different methods is given in Fig.~\ref{fig: exampleresult}.

\renewcommand{\arraystretch}{1.2}
\begin{table}[h!]
    \caption{\label{t:t1}Performance of metric depth estimation methods on opaque objects under different camera poses.}   
    \centering
    \setlength{\arrayrulewidth}{1pt} 
 \begin{tabular}{@{\hspace{4pt}} c @{\hspace{4pt}} c @{\hspace{4pt}} c @{\hspace{4pt}} c @{\hspace{4pt}} c @{\hspace{4pt}} c @{\hspace{4pt}} c @{\hspace{4pt}} c @{\hspace{4pt}} c @{\hspace{4pt}}}
    \hline
&& $\delta_{1.05}\uparrow$ & $\delta_{1.10}\uparrow$ & $\delta_{1.25}\uparrow$&REL$\downarrow$ & RMSE $\downarrow$ & MAE$\downarrow$  \\
\hline
 &\nlra &\textbf{0.8492}&\textbf{0.9939}&\textbf{1.0000}&\textbf{0.0266}&\textbf{0.0164}&\textbf{0.0134}
 \\
\;\;\;Pose 1\;\;\; &\glra & 0.7641&0.9829&1.0000&0.0320&0.0195&0.0163 \\
 &\lwlr &0.6077&0.9426&0.9998&0.0423&0.0249&0.0213 \\
&\damone  &0.0577&0.1803&0.5572&0.2385&0.1231&0.1209  \\
  \hline
    &\nlra & \textbf{0.6384}&\textbf{0.9069}&\textbf{0.9995}&\textbf{0.0446}&\textbf{0.0255}&\textbf{0.0205} \\
Pose 2&\glra &0.1263&0.3002&0.8687&0.1269&0.0695&0.0620 \\
 &\lwlr & 0.2441&0.4201&0.8249&0.1176&0.0673&0.0570 \\
&\damone  &0.1592&0.3114&0.7250&0.1849&0.0902&0.0846\\
  \hline
    &\nlra & \textbf{0.7680}&\textbf{0.9871}&\textbf{1.0000}&\textbf{0.0329}&\textbf{0.0195}&\textbf{0.0161} \\
Pose 3&\glra &0.2759&0.6760&0.9994&0.0733&0.0409&0.0370 \\
 &\lwlr & 0.5622&0.9184&1.0000&0.0453&0.0275&0.0232\\
&\damone  &0.4879&0.7875&0.9840&0.0632&0.0356&0.0315\\
  \hline
 &\nlra & \textbf{0.5766}&\textbf{0.8877}&\textbf{1.0000}&\textbf{0.0468}&\textbf{0.0244}&\textbf{0.0204}\\
Pose 4&\glra &0.2434&0.5212&0.9637&0.0925&0.0455&0.0400\\
 &\lwlr & 0.3739&0.6831&0.9879&0.0723&0.0368&0.0319\\
&\damone  & 0.0327&0.0611&0.2988&0.3342&0.1429&0.1407 \\
  \hline
    \end{tabular}
\end{table}
\renewcommand{\arraystretch}{1.0}

\nlra, our proposed \moma implementation, consistently outperforms other methods for most poses and metrics, showing the highest accuracy and lowest error rates. 
\nlra exhibits the most stable performance across poses. In particular, in all settings, \nlra procured sufficiently high $\delta_{1.10}$ accuracy ($> 0.85$) with MAE $< 0.03$, which we observe as necessary for enabling downstream robotic manipulation tasks. \glra performs well in certain poses but degrades significantly in others. \lwlr shows good performance, often ranking second or third. \damone consistently performs poorly across all poses.
The results suggest that \nlra is the most effective method for depth estimation of opaque objects across various poses, while \damone requires significant improvements to be competitive in this task.

\subsection{Ablation Study: Impact of Ground Truth Samples}\label{sec:n}
\begin{figure}[htb]
    \centering
    \includegraphics[width=1\linewidth]{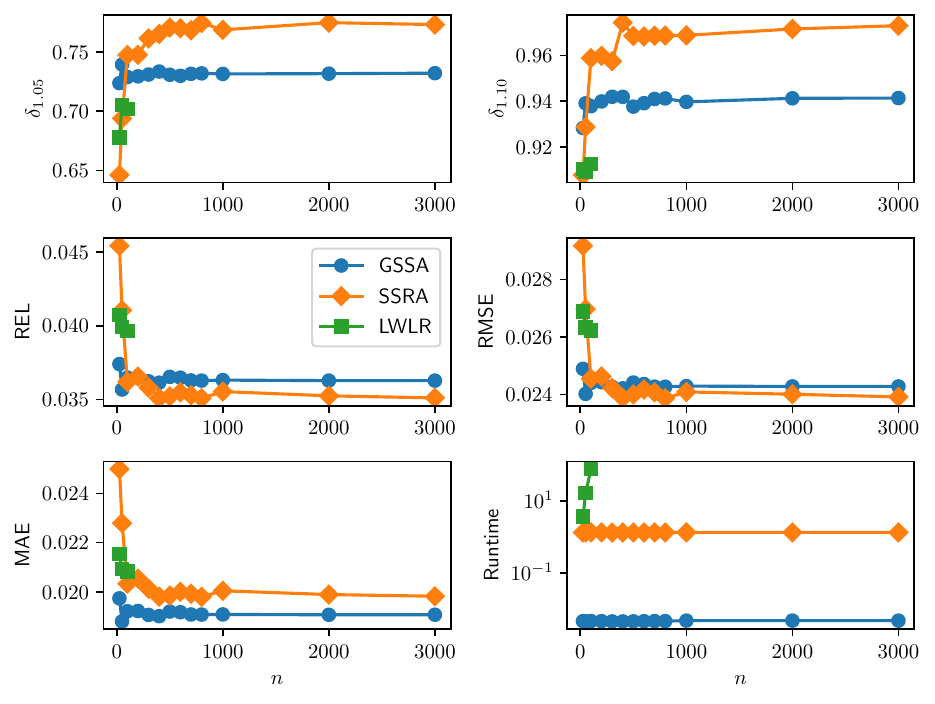}
    \caption{Impact of the number of ground truth depth points used in performing that alignment (camera Pose 1 as shown in Fig.~\ref{fig:pose_study}).
    }
    \label{fig:samples}
\end{figure}

To evaluate the impact of ground truth samples, for camera Pose 1, we vary the number of randomly selected ground truth depth samples ($n$) from 20 to 3000. The results, plotted in Fig.~\ref{fig:samples}, indicate that good metric depth can be obtained with only tens to hundreds of samples. In particular, it is not meaningful to use more than $3$-$400$ ground truth samples. \nlra outperformed \glra and \lwlr in terms of metrics. Additionally, \nlra demonstrated significantly reduced runtime compared to \lwlr, owing to its fewer parameters that require optimization.

\subsection{Ablation Study:  Impact of Normalization}
In this experiment, we examine the impact of the depth-normalization. Tab.~\ref{t:3} provides the performance of various normalization methods evaluated on the dataset for camera Pose 1. 
It is readily clear that the min-max normalization method achieves the highest performance across all three alignment methods. Specifically, \nlra with min-max normalization attains an MAE of 0.0134 and a $\delta_{1.05}$ of 0.8492. In comparison, the median normalization method, while less effective than min-max normalization, outperforms the method that employs no normalization techniques.

\renewcommand{\arraystretch}{1.2}
\begin{table}[h!]
    \caption{Impact different normalization methods (camera Pose 1).}    
    \label{t:3}
    \centering
    \setlength{\arrayrulewidth}{1pt} 
   \begin{tabular}{@{\hspace{4pt}} c @{\hspace{4pt}} c @{\hspace{4pt}} c @{\hspace{4pt}} c @{\hspace{4pt}} c @{\hspace{4pt}} c @{\hspace{4pt}} c @{\hspace{4pt}} c @{\hspace{4pt}} c @{\hspace{4pt}}}
    \hline
&& $\delta_{1.05}\uparrow$ & $\delta_{1.10}\uparrow$ & $\delta_{1.25}\uparrow$&REL$\downarrow$ & RMSE $\downarrow$ & MAE$\downarrow$ \\
  \hline
 &\nlra &\textbf{0.8492}&\textbf{0.9939}&\textbf{1.0000}&\textbf{0.0266}&\textbf{0.0164}&\textbf{0.0134} \\
\;\;Min-Max\;\;&\glra & 0.7641&0.9829&1.0000&0.0320&0.0195&0.0163 \\
 &\lwlr &0.6077&0.9426&0.9998&0.0423&0.0249&0.0213 \\

  \hline
    &\nlra & 0.5473&\textbf{0.8463}&\textbf{0.9988}&\textbf{0.0507}&\textbf{0.0305}&0.0254 \\
Median&\glra &\textbf{0.5595}&0.8415&0.9973&0.0509&0.0308&\textbf{0.0253} \\
 &\lwlr & 0.4480&0.7247&0.9773&0.0671&0.0403&0.0330 \\
  \hline
   &\nlra & \textbf{0.3658}&\textbf{0.6582}&\textbf{0.9929}&\textbf{0.0812}&\textbf{0.0443}&\textbf{0.0413} \\

None &\glra &0.2321&0.4395&0.8972&0.1193&0.0648&0.0607 \\
 &\lwlr & 0.2280&0.4351&0.8805&0.1232&0.0673&0.0626\\
&\damone  &0.0577&0.1803&0.5572&0.2385&0.1231&0.1209  \\
  \hline
    \end{tabular}
\end{table}
\renewcommand{\arraystretch}{1.0}

On the other hand, in the absence of normalization (denoted as \texttt{None}), the performance across all metrics significantly deteriorates for all methods. As discussed in Sec.~\ref{sec:algorithm}, without normalization, the predicted depth range from \mdem experiences substantial fluctuations, rendering the calibration ineffective.

\subsection{Ablation Study: Impact of \mdem Models}
To select the best \mdem models, in addition to \damone, we evaluated \damtwo~\cite{yang2024depth} and \metricddd~\cite{hu2024metric3d}, which achieve impressive performance on several existing public metric depth estimation benchmarks. Whereas \damone\ is fine-tuned on public transparent object datasets, we directly use the pre-trained checkpoints for \metricddd\ and \damtwo. The results are shown in Tab.~\ref{t:different_models}. The method with/without ``+'' indicates whether the \nlra\ alignment module from Subsection~\ref{subsec:poses} is used/not used. We observe the raw metric depth predictions from these \mdem\ models are far from precise enough for direct use in robotic manipulation. However, with \moma's \nlra alignment, the mean absolute error (MAE) drops to $0.01 - 0.03$, suitable for robotic manipulation. We found \damone fine-tuned on a mixed transparent object dataset achieves better performance. This is due to the datasets sharing more similarities in the depth distribution pattern from the experimental setup we used for robotic manipulation.

\renewcommand{\arraystretch}{1.2}
\begin{table}[h!]
    \caption{Metrics Evaluation, different \mdem models.}    
    \label{t:different_models}
    \centering
    \setlength{\arrayrulewidth}{1pt} 
    \begin{tabular}{@{\hspace{4pt}} c @{\hspace{4pt}} c @{\hspace{4pt}} c @{\hspace{4pt}} c @{\hspace{4pt}} c @{\hspace{4pt}} c @{\hspace{4pt}} c @{\hspace{4pt}} c @{\hspace{4pt}} c @{\hspace{4pt}}}
    \hline
&& $\delta_{1.05}\uparrow$ & $\delta_{1.10}\uparrow$ & $\delta_{1.25}\uparrow$&REL$\downarrow$ & RMSE $\downarrow$ & MAE$\downarrow$ \\
  \hline
  &\damone &0.0577&0.1803&0.5572&0.2385&0.1231&0.1209 \\
  &\damone{}+ &\textbf{0.8492}&\textbf{0.9939}&\textbf{1.0000}&\textbf{0.0266}&\textbf{0.0164}&\textbf{0.0134} \\

  &\damtwo & 0.0000&0.0000&0.0000&1.4410&0.7396&0.7356 \\
Pose 1&\damtwo{}+ & 0.8030&0.9695&0.9999&0.0302&0.0188&0.0156 \\
 &\metricddd &0.0000&0.0000&0.0022&0.6378&0.3279&0.3251 \\
&\metricddd{}+ &0.7876&0.9814&0.9983&0.0335&0.0198&0.0168 \\
  \hline
&\damone &0.0540&0.2637&0.8566&0.1619&0.1003&0.0948\\
  &\damone{}+ &\textbf{0.7290}&\textbf{0.9178}&\textbf{0.9910}&\textbf{0.0395}&\textbf{0.0349}&\textbf{0.0234} \\

  &\damtwo & 0.0002&0.0003&0.0020&1.1547&0.7181&0.7050 \\
Standing&\damtwo{}+ & 0.4717&0.7487&0.9602&0.0703&0.0586&0.0437 \\
 &\metricddd &0.0038&0.0097&0.0966&0.4532&0.2894&0.2788 \\
&\metricddd{}+ &0.2766&0.4600&0.6242&0.1928&0.1307&0.1157 \\
    \hline
&\damone &0.2057&0.4252&0.8596&0.1288&0.0641&0.0616 \\
  &\damone{}+ &0.5708&0.8547&0.9973&0.0494&0.0280&0.0242 \\
  &\damtwo & 0.0000&0.0000&0.0000&2.3565&1.2428&1.1237\\
In Tote&\damtwo{}+ &0.4693&0.8684&1.0000&0.0553&0.0314&0.0267 \\
 &\metricddd &0.0000&0.0000&0.0062&0.5121&0.2460&0.2445 \\
&\metricddd{}+ &\textbf{0.6212}&\textbf{0.8886}&\textbf{0.9996}&\textbf{0.0470}&\textbf{0.0260}&\textbf{0.0223} \\

  \hline
    \end{tabular}
\end{table}
\renewcommand{\arraystretch}{1.0}

\subsection{Transparent Objects}
\begin{figure}[t!]
    \centering
    \includegraphics[width=1\linewidth]{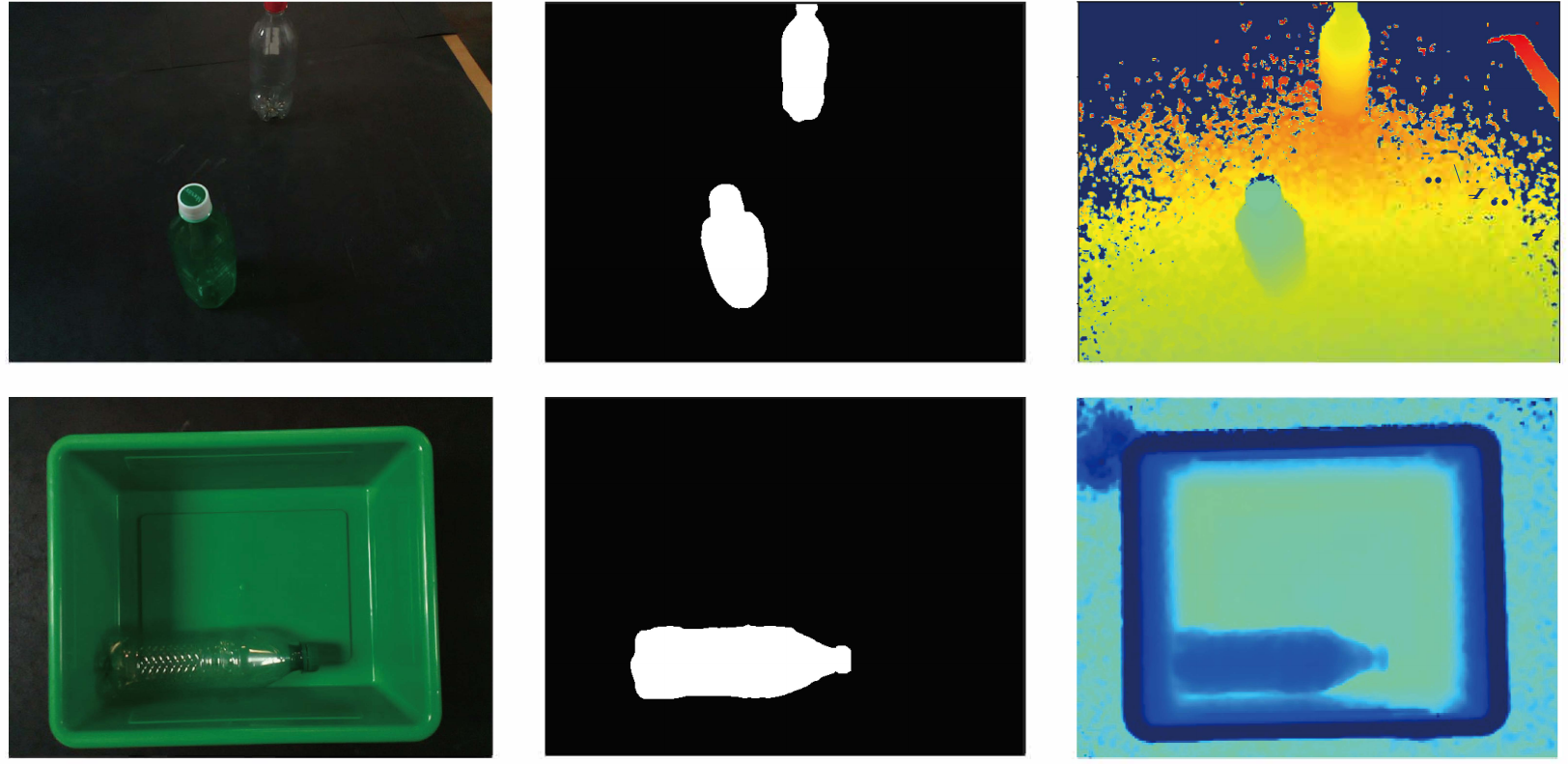}
    \caption{[top] An example containing RGB, mask, and ground-truth depth of the transparent object dataset used in the evaluation for the two-finger grasping downstream task. [bottom] The same for the suction-based bin-picking downstream task.}
    \label{fig:transparent_objects}
\end{figure}

We used the L515 sensor to collect a testing dataset comprising 10 transparent objects. For each object, 100 images were captured in two scenarios with fixed camera poses: one with the objects standing on a table and another with the objects lying in a tote. To obtain pseudo-ground-truth depth for the transparent objects, we replaced each transparent object with an opaque object of the same shape and position. Sample data is given in Fig.~\ref{fig:transparent_objects}. 

The evaluation results are summarized in Tab.~\ref{t: trans}. \nlra achieved a $\delta_{1.05}$ score of 0.73 for standing transparent objects and 0.57 for objects in the tote. The lower performance in the tote scenario may be attributed to the limited similarity between the fine-tuning dataset and the tote environment. Nevertheless, the $\delta_{1.05}$ score is above $0.85$ and the mean absolute error (MAE) for the tote scenario is 0.0242, sufficient for downstream suction-based picking.

\renewcommand{\arraystretch}{1.2}
\begin{table}[h!]
    \caption{Performance on transparent objects.}    
    \label{t: trans}
    \centering
    \setlength{\arrayrulewidth}{1pt} 
      \begin{tabular}{@{\hspace{4pt}} c @{\hspace{4pt}} c @{\hspace{4pt}} c @{\hspace{4pt}} c @{\hspace{4pt}} c @{\hspace{4pt}} c @{\hspace{4pt}} c @{\hspace{4pt}} c @{\hspace{4pt}} c @{\hspace{4pt}}}
    \hline
&& $\delta_{1.05}\uparrow$ & $\delta_{1.10}\uparrow$ & $\delta_{1.25}\uparrow$&REL$\downarrow$ & RMSE $\downarrow$ & MAE$\downarrow$ \\
  \hline
&\nlra &\textbf{0.7290}&\textbf{0.9178}&\textbf{0.9910}&\textbf{0.0395}&\textbf{0.0349}&\textbf{0.0234}\\
\;\;\;Standing\;\;\; &\glra &0.5806&0.7709&0.9904&0.0582&0.0453&0.0340 \\
 &\lwlr &0.5964&0.7828&0.9904&0.0561&0.0445&0.0329 \\
&\damone  &0.0540&0.2637&0.8566&0.1619&0.1003&0.0948  \\
  \hline
 &\nlra &\textbf{0.5708}&\textbf{0.8547}&\textbf{0.9973}&\textbf{0.0494}&\textbf{0.0280}&\textbf{0.0242} \\

In Tote &\glra &0.5640&0.8548&0.9973&0.0495&0.0280&0.0242 \\
 &\lwlr &0.4289&0.7866&0.9957&0.0627&0.0341&0.0306 \\
&\damone  &0.2057&0.4252&0.8596&0.1288&0.0641&0.0616  \\
\hline
    \end{tabular}
\end{table}
\renewcommand{\arraystretch}{1.0}

\subsection{Real Robot Experiments}
\begin{figure}[tb!]\label{fig:objects}
    \centering
    \vspace{2mm}
\includegraphics[width=\linewidth]{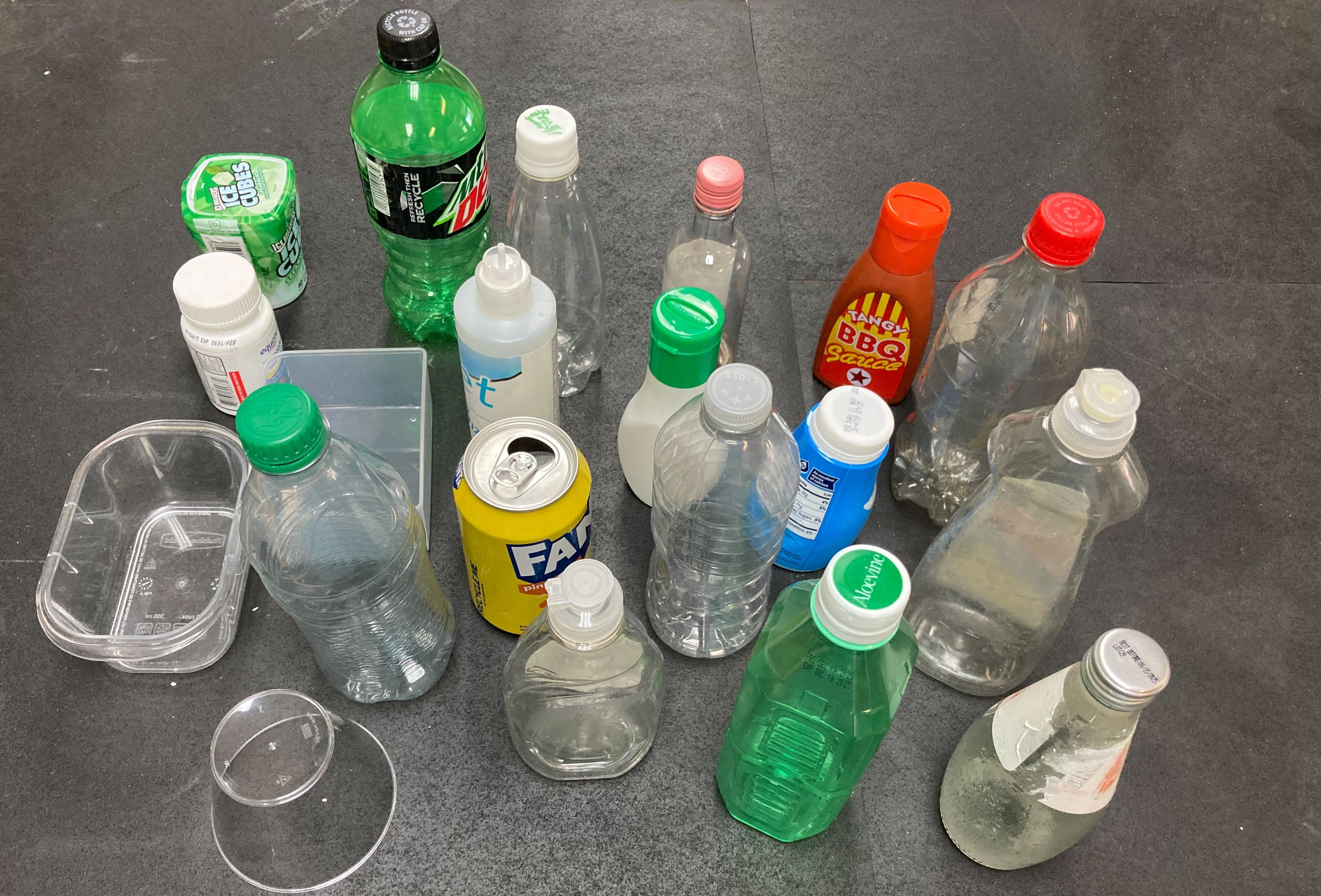}
    \vspace{1mm}
    \caption{Objects tested in the real robot experiment. 
    }
    \vspace{1mm}
    \label{fig:tested_objects}
\end{figure}

As downstream applications, we conducted real robot experiments using a UR-5e robot equipped with an adaptive Robotiq  2f-85 gripper for finger-based grasping and a vacuum gripper for suction-based grasping. The transparent and opaque testing objects are depicted in Fig.~\ref{fig:tested_objects}. The adaptive gripper was used to grasp objects lying or standing on the table, while the vacuum gripper was employed for suctioning objects placed in the green tote.

For each scenario, the camera was fixed, and calibration was performed using a single RGBD image captured by the L515 sensor. After the calibration, the depth sensor was disabled for methods utilizing \mdem. We conducted 50 trials for each scenario, randomly placing 2-3 objects each time, to determine the success rate of each method. The results are presented in Tab.~\ref{t: success_rate}.
\nlra achieves fairly decent success rates of over $80\%$ on opaque objects, beating other alignment methods, especially in the two-finger grasp scenario. This is partially due to suction-based grasping having built-in compliance, e.g., most grasps are top-down; overshooting a little will still work.  
As a reference, using full RGBD information acquired from the L515 sensor, all opaque objects can be grasped (either fingered or suction-based) whereas no transparent objects can be grasped using the raw depth from the L515 sensor. 
\jy{A question that I have is whether we can use ground truth depth from more than one image to calibrate.}

\renewcommand{\arraystretch}{1.2}
\begin{table}[h!]
    \caption{Success rates in downstream manipulation tasks. }    
    \label{t: success_rate}
    \centering
    \footnotesize
    \setlength{\arrayrulewidth}{1pt} 
    \begin{tabular}{l c c c c c c c }
    \hline
 & \nlra& \glra& \lwlr \\
  \hline
Two-finger grasp: opaque objects & $82\%$ &$72\%$ &$62\%$ \\
Two-finger grasp: transparent objects &$70\%$  & $56\%$  & $52\%$\\
  \hline
Suction grasp: opaque objects& $86\%$ & $86\%$ &$80\%$ \\
Suction grasp: transparent objects&$78\%$ & $76\%$&$70\%$ \\
\hline
    \end{tabular}
\end{table}
\renewcommand{\arraystretch}{1.0}

%% file: texs/07-conclusion.tex
\jy{My edit location}
Leveraging the power of vision foundational models, in this work, we propose a \emph{monocular one-shot metric-depth alignment} (\moma) framework toward enabling depth-sensor-free estimation of metric object depth for executing robotic manipulation tasks.
As a first attempt at developing the capability to perform robotic manipulation without costly and sometimes unreliable depth sensors, \moma holds significant application potential, especially in industrial settings where cameras and robot arms are fixed.
Our specific \moma implementation, \nlra, running very fast at both alignment parameter calibration (few seconds) and inference time (few milliseconds, achieves significantly better accuracy in standard benchmark tests (e.g., $\delta_{1.05}$, $\delta_{1.10}$, RMSE, MAE) when compared with similar state-of-the-art methods. Moreover, in downstream robotic two-finger and suction-based grasping tasks, \nlra consistently achieves over $80\%$ success rates when handling non-transparent objects and over $70\%$ success rates on transparent objects. 

While \moma $+$ \nlra shows great promise in enabling monocular and RGB only metric depth estimation, it represents an initial step in rendering such attempts fully practical. As of now, \moma still has some catch up to do when compared with RGB-D based methods, especially when it comes to dealing with transparent objects \cite{sajjan2020clear,shi2024asgrasp}. 
With that in mind, future research is currently underway to further boost the performance/robustness of \moma to realize its full potential. We note that \nlra only uses depth data from a single scene for calibration; using multiple scenes (i.e., going from one shot to few shots), even without increasing $n$, the total number of points could improve the alignment accuracy due to the alignment data being more diverse. Without the need for expensive depth sensors or dedicated on-board stereo computation, stereo or multi-view versions of \moma could be developed as well, which will provide more precise depth estimates.  